\documentclass[conference]{IEEEtran}
\IEEEoverridecommandlockouts
\usepackage{cite}
\usepackage{amsmath,amssymb,amsfonts}
\usepackage{algorithmic}
\usepackage{graphicx}
\usepackage{textcomp}
\usepackage{xcolor}

\usepackage{graphics} % for pdf, bitmapped graphics files
\usepackage{epsfig} % for postscript graphics files
\usepackage{mathptmx} % assumes new font selection scheme installed
\usepackage{times} % assumes new font selection scheme installed
\usepackage{svg}
\usepackage{multirow}
\usepackage{url}
\usepackage{hyperref}

\def\BibTeX{{\rm B\kern-.05em{\sc i\kern-.025em b}\kern-.08em
    T\kern-.1667em\lower.7ex\hbox{E}\kern-.125emX}}

\title{SynthoGestures: A Multi-Camera Framework for Generating Synthetic Dynamic Hand Gestures for Enhanced Vehicle Interaction
}

\author{Amr Gomaa$^{*}$, Robin Zitt$^{*}$, Guillermo Reyes$^{*}$ and Antonio Kr{\"u}ger$^{*}$% <-this % stops a space
\thanks{$^{*}$All authors are with the German Research Center for Artificial Intelligence (DFKI) and Saarland Informatics Campus, Saarland University, 66123 Saarbrucken, Germany.
        {Corresponding author: \tt\small amr.gomaa@dfki.de} 
        Author Version.}}

\IEEEpubid{\makebox[\columnwidth]{979-8-3503-4881-1/24/\$31.00~\copyright~2024 IEEE \hfill} \hspace{\columnsep}\makebox[\columnwidth]{ }}
        
\begin{document}

\maketitle

\begin{abstract}

    Dynamic hand gesture recognition is crucial for human-machine interfaces in the automotive domain. However, creating a diverse and comprehensive dataset of hand gestures can be challenging and time-consuming, especially in dynamic dual-task situations like driving. To address these challenges, we propose using synthetic gesture datasets generated by virtual 3D models as an alternative. Our framework synthesizes realistic hand gestures using a combination of 3D models and animation software, particularly utilizing Unreal Engine. This approach enables the creation of diverse and customizable gesture datasets, reducing the risk of overfitting and improving the model's generalizability. Specifically, our framework generates natural-looking dynamic hand gestures with multiple variants, including gesture speed, performance, and hand shape.
    Moreover, we simulate various camera locations, such as above the driver and behind the wheel, and different camera types, such as RGB, infrared, and depth cameras, without incurring additional time and cost to obtain these cameras. Our experiments demonstrate that our proposed framework, SynthoGestures (available at https://github.com/amrgomaaelhady/SynthoGestures), can augment or replace existing real-hand datasets with additional enhancement in gesture recognition accuracy. Our tool for generating synthetic static and dynamic hand gestures saves time and effort in creating large datasets, facilitating the faster development of gesture recognition systems for automotive applications.
\end{abstract}

%%%%%%%%%%%%%%%%%%%%%%%%%%%%%%%%%%%%%%%%%%%%%%%%%%%%%%%%%%%%%%%%%%%%%%%%%%%%%%%%
\section{Introduction}

Hand gestures are an essential aspect of human-machine interaction, as they enable natural communication between humans and devices~\cite{jaramillo2020real,panwar2011hand, zhu2018control,abner2015gesture,bellugi1972comparison,dardas2011hand}. In particular, hand gesture recognition can control various parts of the vehicle, such as audio systems, climate control, and navigation~\cite{pickering2007research, young2020designing, zengeler2018hand}. Hand gesture recognition involves detecting, analyzing, and interpreting hand movements to understand the intended message~\cite{tsai2020design, zhu2018control,bressem2011rethinking,molchanov2016online,kopuklu2020online,kopuklu2019real}. As technology advances rapidly, hand gesture recognition techniques have evolved significantly. Despite these advances, state-of-the-art techniques rely on complex deep learning algorithms that require large amounts of data~\cite{jaramillo2020real, kao2011human, panwar2011hand, shanthakumar2020design}. While most existing state-of-the-art gesture recognition methods achieve high performance, they suffer from a lack of generalization due to multiple factors, such as the need for further data and dataset biases. To solve these problems, researchers used model adaptation and personalization to improve user performance~\cite{junokas2018enhancing, nelson2015adaptive, lou2017personalized}; however, they still require recruiting, collecting and recording large amounts of data. Obtaining large datasets of real hand gestures in dual-task dynamic situations, such as controlling the vehicle while driving, can be expensive and time-consuming. To address this challenge, researchers have turned to synthetic data generation~\cite{zhao2022low, wang2018research,roberto2017procedural,lindgren2018learned,tsai2017synthetic,Ibrahim2013VisualSD,kanis2018improvements,memo2015exploiting}; however, these tools are still generating basic and mostly static gestures. Thus, we propose \textit{SynthoGestures}\footnote{\url{https://github.com/amrgomaaelhady/SynthoGestures}}, an Unreal Engine-based framework that creates synthetic nature-looking dynamic hand gestures with all the desired variations to be used in training large gesture recognition models. Our tool can be used to create synthetic datasets from multiple camera positions within the vehicle (i.e., multiple point-of-view) as well as multiple sensor types, such as RGB, infrared, and depth-based cameras, with added noise modeled from existing hardware. 

\begin{figure}[t]
    \centering
    \includegraphics[width=\linewidth]{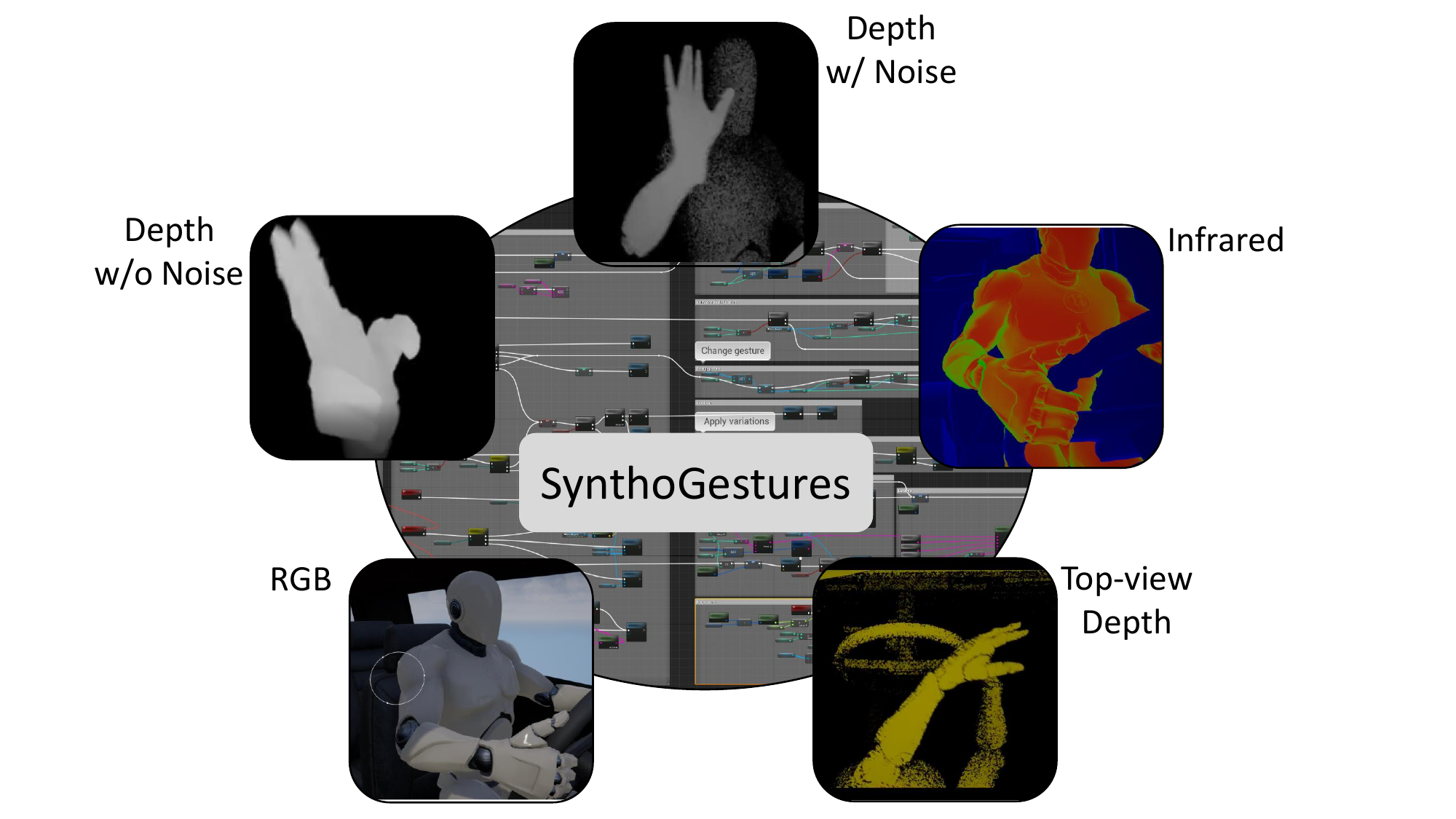}
    \caption{We present, a new 3D synthetic hand gesture generation framework, \textit{SynthoGestures}, that provides a cost-effective and flexible approach for creating new variations of dynamic and static hand gestures. Our framework combines 3D modeling with a game engine (i.e., Unreal Engine) to produce multiple datasets with different camera positions (e.g., infotainment perspective, top view, and behind the wheel) and different camera types (e.g., RGB, infrared, and depth camera) with different noise modeling techniques.}
    \label{fig:teaser}
\end{figure}

In summary, our \textbf{contributions} are three folds as follows: 1) We present, \textit{SynthoGestures}, a novel 3D dynamic hand gesture generation framework that provides a cost-effective and flexible approach for creating new variations of hand gestures using different sensor types to overcome data set biases and lack of generalization. 2) We provide a dataset of synthetic hand gestures in the automotive domain for controlling various parts of the infotainment system. 3) We highlight the augmentation ability of synthetically generated dynamic hand gestures to partially or fully replace existing hand gesture datasets with additional enhancement in model performance when training with state-of-the-art recognition models.

\section{Related work}

This section delves into existing research on synthetic gesture generation and recognition. When utilizing synthetic data to train a machine learning algorithm for gesture recognition, two overarching aspects emerge; the quality and quantity of the samples. Hence, synthetic data generation is challenged by generating a large volume of samples that closely resemble real gestures in terms of appearance and functionality while encompassing sufficient variations to generate a substantial amount of data that would enhance generalization and model performance.

Several approaches explored synthesizing hand gestures to enhance gesture recognition models~\cite{lindgren2018learned,tsai2017synthetic,Ibrahim2013VisualSD,fowley2021sign,kim2019improving,taranta2016rapid,de2020vision}. However, most approaches focused on static, isolated hand gesture recognition with a black background. More specifically, Lindgren et al.~\cite{lindgren2018learned} investigated simple static gesture recognition using synthetic data. However, they focused on the impact of simple variations in the simulated hand (e.g., thicker hands, taller fingers, and wider interfinger spacing) on recognition accuracy. In more detail, they used a character model in Unreal Engine to perform five static hand gestures. While the length and spacing of the fingers were modified within the 3D model, the remaining variations, such as the thickness of the fingers, the rotation, and the position of the hand, were applied to the depth image after generation. They evaluated the influence on recognition accuracy by systematically excluding these individual variations from the training set in an ablation study. The results revealed that all variations, except rotation, contributed to increased recognition accuracy. Interestingly, the accuracy declined when rotation changes were included in the training set. The authors reasoned that the original orientation of the hand played a crucial role in recognition and that alterations in rotation disrupted this essential feature, rendering it unreliable for accurate classification. However, a limitation of this study is that the authors primarily focused on applying variations to the image itself rather than leveraging the full potential of the character model. Consequently, their approach could have been replicated using real hand images, thereby underutilizing the capabilities offered by the game engine. Another limitation is that they focused only on static gestures in a static environment, unlike our approach, which focuses on dynamic gestures for a dynamic environment such as driving. 

\begin{figure*}[t]
\centering
\includegraphics[width=\textwidth]{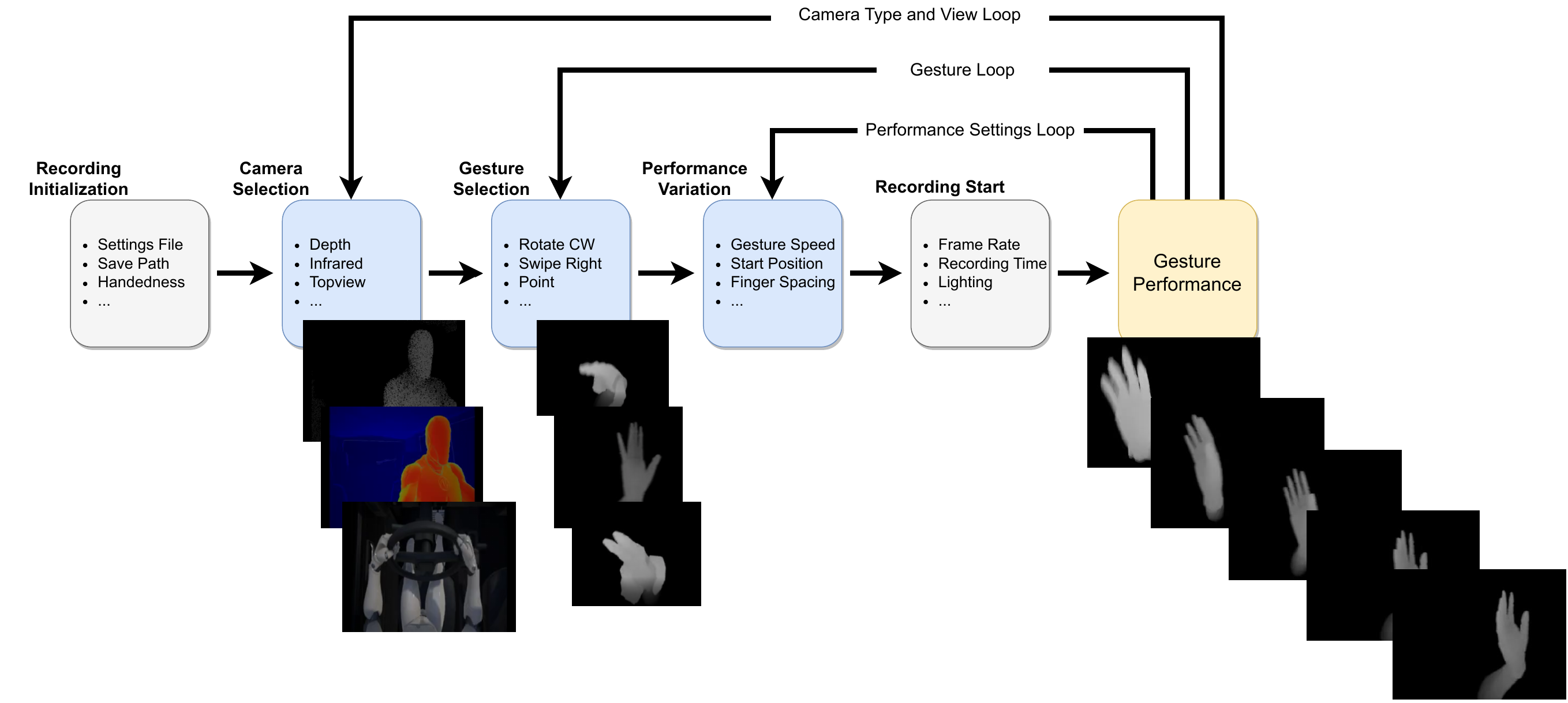}
\caption{
\textbf{Overview of our SynthoGestures framework} for dynamic hand gesture synthesis. It takes initial settings by user input or reads them from a JSON file. This includes generic recording settings such as saving file path, number of recordings per gesture, video resolution, and default hand to perform gestures. Next, the framework allows for the customization of gestures, including camera types, locations, hand shapes and positions, gesture speed, and lighting conditions. This enables the generation of diverse variations in performance for dynamic hand gestures. Additionally, it allows the user to identify only general settings and automatically loops over all possible gesture-specific settings to produce a comprehensive data set with multiple variations.
} 
\label{fig:overview}
\end{figure*}

Similarly, Tsai et al.~\cite{tsai2017synthetic} explored an alternative approach to improve recognition accuracy by combining synthetic and real data. Their methodology aligned with the previous study, involving variations in static hand gestures that encompass finger spacing, rotation, size, and position. However, they implemented these modifications directly on a hand model, incorporating a more comprehensive range of gradations. This resulted in a total of 108,864 synthetic gestures generated.
They compared three training models: one that exclusively uses synthetic data, another that is trained sequentially with synthetic data followed by real data, and a third model that is trained simultaneously with synthetic and real data. The authors reported an accuracy improvement of 27.78\%, 79.86\%, and 89.58\% for each model, respectively. However, when the authors introduced a complex background instead of a simple black background, all models fell significantly below the accuracy value of the real data-only model. While the authors suggested employing a background removal algorithm to mitigate this issue, we suggest incorporating the background during the gesture modeling and synthesis stage, as presented in our simulation. 
Although the aforementioned approach involves augmenting real data with synthetic data, most studies adopt a slightly different strategy. They employ a real data set as the foundation and supplement it with synthetic data to expand the training set. This approach reduces the need for extensive gesture recording while enabling coverage of corner cases not covered by the real data set. For example, Ibrahim and Kashef~\cite{Ibrahim2013VisualSD} observed that dynamic sign language gesture recognition improved consistently when using a combination of real and synthetic data, especially for small-size datasets. Additionally, they discovered that synthetic data promote signer independence within the training set, mitigating the problem of overfitting and improving model generalization by ensuring sufficient variation in gesture generation.

The significance of recognition accuracy in gesture recognition extends beyond animation quality and realistic samples. De Melo et al.~\cite{de2020vision} demonstrated this by generating dynamic gestures using 3D character models and investigating the performance impact of various variations in the virtual environment and gestures. Experimental manipulation of input parameters yielded notable results. Generally, higher resolution proved beneficial for performance. Similar trends were observed for rendering quality and frame rate, as reducing the number of frames resulted in diminished performance.
Additional experiments were conducted to assess the influence of factors such as animation speed, skin tone, thickness, character model gender, and background environment. Any modifications to these variations led to a significant decrease in performance, except for the background and gender of the character model. The findings of this study highlight the multitude of factors that can impact the performance of gesture recognition algorithms. Consequently, these factors warrant careful consideration when synthesizing hand gestures and justify the need for a framework with more control over the generated data as implemented in our approach.

\section{Method}

The \textit{SynthoGestures} framework employs a systematic approach to generate dynamic hand gestures. It starts by collecting initial settings from user input or a JSON file, including recording parameters and default hand selection. Users can customize gestures by specifying camera types, locations, hand attributes, gesture speed, and lighting conditions. The framework automates the generation process by iterating through gesture-specific settings, resulting in a comprehensive dataset with various performance variations, as seen in~\autoref{fig:overview}.
An initial gesture script (i.e., Unreal Engine blueprint) is created as a baseline for further gesture generation. It encompasses a list of variations that lead to different executions during generation and recording. Once the gestures are defined, the user proceeds with camera location selection, choosing a camera type and angle for iterative recording.

\paragraph{Camera Selection.} 
Users can customize camera setups in camera settings, including camera types and their associated parameters. Each camera type in the list includes parameters such as activation status and a list of camera angles such as position and rotation. Within our framework, three camera types are available: RGB, depth, and infrared. However, additional camera types can be easily modeled into the framework using the appropriate realistic specifications. The RGB image is obtained directly from the virtual camera sensor provided by Unreal Engine without additional post-processing or added effects. The depth camera introduces depth perception by utilizing a flip-book animation of a grayscale noise texture. This texture, divided into multiple tiles, changes every few frames to simulate noise variations. An adjustable variable controls the intensity of the noise, filtering out pixels with smaller alpha values. The noise intensity depends on the object's distance from the camera and the depth difference between adjacent pixels. This approach enhances the noise around edges and nonorthogonal surfaces. The resulting depth image is generated by interpolating pixel colors based on the distance to the camera, while the depth difference and distance modulate the noise. This approach for modeling the noise of a depth camera provides a realistic representation of depth information in the scene as seen in previous work~\cite{sweeney2019supervised}.
The infrared effect is achieved by applying the Fresnel effect~\cite{wang2019effect} to the person's body, resulting in an orange color at the center, transitioning to green towards the edges. Other objects, including the car, exhibit the Fresnel effect with a dark blue appearance. Blurring effects are simulated using a panned noise texture across the character, creating a blurry appearance at the edges.

\begin{figure*}[t]
\centering
\includegraphics[width=0.7\textwidth]{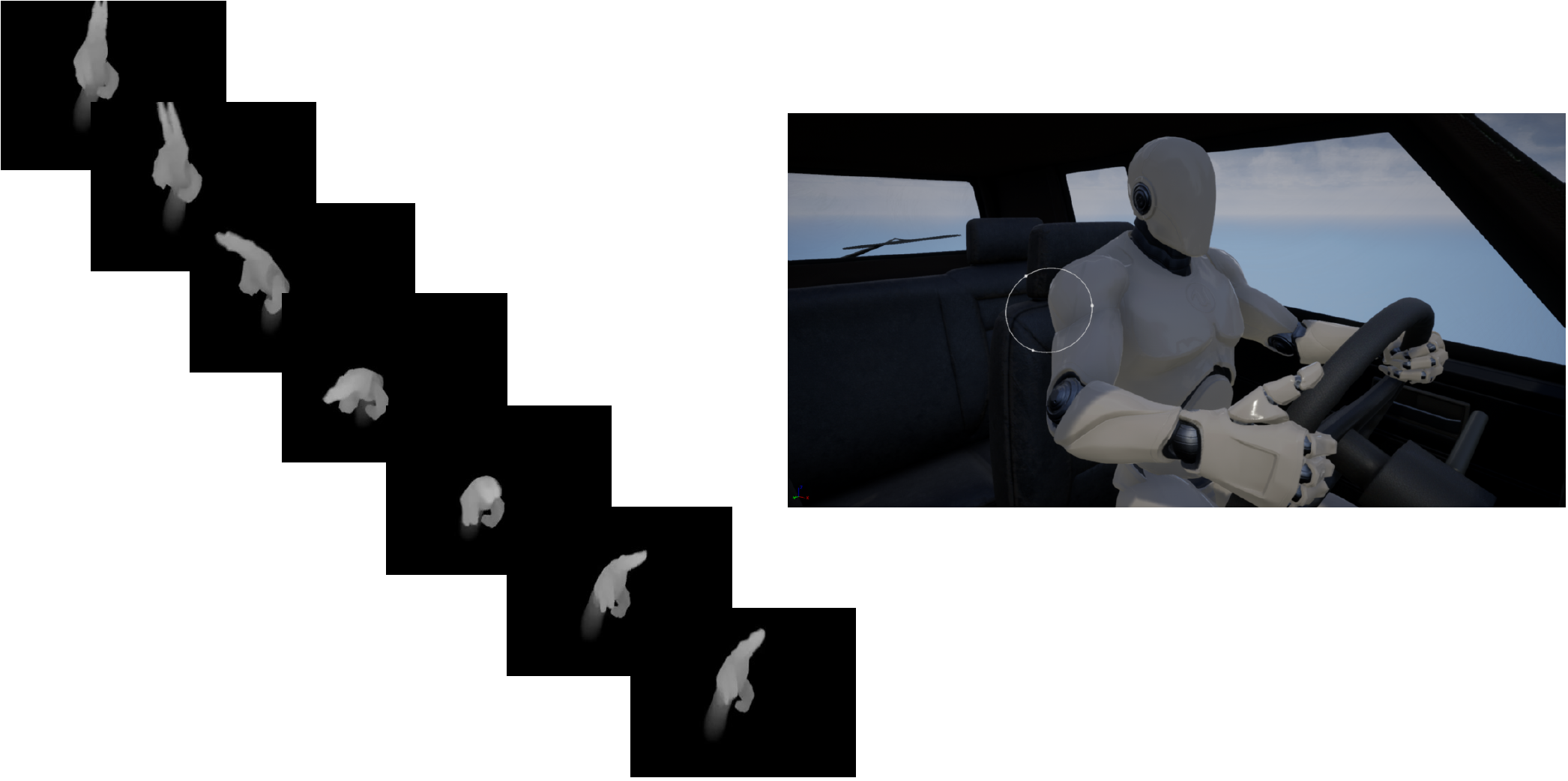}
\caption{An example of the character model with the spline of a rotation gesture (right) with the generated gesture sequence as a low frame rate depth image without noise (left).} 
\label{fig:rotationgesture}
\end{figure*}

\paragraph{Gesture Selection and Performance Variation.} 
Upon selecting the desired camera angle, the system proceeds with the iteration process through the available gestures. The editable gesture list found in the general settings determines the inclusion of gestures in the generation phase. A child script inherent from the previously mentioned baseline script to provide essential gesture variations and to define the hand's movement path during the gesture execution using a spline positioned in front of the human model within arm's reach. An illustrative example of setting up a rotation gesture is presented in~\autoref{fig:rotationgesture}.
Our gesture generation system encompasses two modes: description-based and animation-based. The description-based method provides a more optimum solution for gesture execution. There are two modes within the description-based method: the single gesture mode and the chain gesture mode.
In the single gesture mode, a pre-gesture is initiated, followed by the execution of the gesture itself, and concluding with a post-gesture. This sequence is repeated to generate multiple variants of the gesture. The same procedure is applied to the remaining gestures. On the other hand, in the chain gesture mode, the order differs. The pre-gesture is initiated first, followed by the execution of all gestures, including transitions between them, and ultimately the post-gesture. This iterative process generates diverse variants. In chain mode, consecutive gestures are considered cohesive and recorded in a single video. Several parameters can be chosen for the gesture, including \textit{Gesture Name}, \textit{Number Of Variations}, \textit{Use of Left Hand}, \textit{Arm Part} (identifying the segment of the arm), \textit{Gesture Speed}, \textit{Pre-gesture Speed} (denoting the speed of the hand when reaching the starting position of the gesture) and \textit{ Post-gesture speed} (denoting the speed of the hand when moving back to the driving wheel). Several additional variations exist within the \textit{SynthoGestures} framework controlling every aspect of gesture performance and hand physique.

\paragraph{Gesture Execution.}
Once all cameras are selected and all gestures are implemented, gesture execution begins. In the description-based approach, a gesture is represented in the game engine as an object comprising a spline and variation components. During recording, the arm follows a coordinate traveling along the spline.
The child script handles the coordination of the spline and arm movement. The main event, triggered after the pre-gesture animation, updates the spline to encompass the entire gesture path. A timeline moves the hand along the spline, while an additional function determines the hand's position and applies appropriate rotations and finger variations.
There are special cases for chain mode and static gestures, each requiring specific handling during gesture execution. Transition speeds are adjusted on the basis of the length of the spline segment and desired speed, ensuring realistic movement.
The \textit{Control Rig}, an inverse kinematic system for human movement modeling, updates the human model based on the position of the hand. It aligns the upper and lower arm with the wrist to achieve natural movement. The arm is positioned before the main gesture for the hand and finger gestures, and only the hand or finger is moved using an aiming function. The Control Rig's functionalities include updating the arm, hand rotation, finger rotation, and spacing, contributing to the realistic execution of the gesture.

\section{Results}

To assess the effectiveness of the gesture generation system, we employ the methodologies of Tsai et al.~\cite{tsai2017synthetic} and Ibrahim and Kashef~\cite{Ibrahim2013VisualSD} in combining synthetic and real data to improve the accuracy of recognition or maintain performance with limited real data points. We utilize the NVIDIA Dynamic Hand Gesture Dataset~\cite{molchanov2016online} and the state-of-the-art gesture recognition model recently developed by K{\"o}p{\"u}kl{\"u} et al.~\cite{kopuklu2019real,kopuklu2020online}. Six distinct gestures were selected from the NVIDIA dataset, generated with multiple variations, and then recorded as depth camera videos. These variations encompassed changes in the character model's gesture speed, position, finger rotations, and hand orientations. The selection of random value ranges for these variations was aimed at producing natural-looking gestures while maximizing the parameter space. In total, 600 gesture videos were generated, with 100 variants for each of the six selected gestures. The real data set was divided into 50\% for training, 30\% for validation, and 20\% for testing. Note that the synthetic data was not used for validation and testing since it is used as an augmentation technique and it is not the target of the gesture recognition model. The six gestures chosen for the experiments included a horizontal swipe gesture, a vertical swipe gesture, a horizontal swipe gesture with two fingers, a peace sign, a rotating gesture with two fingers, and a pointing gesture with two fingers.

\begin{figure}[t]
\centering
\includegraphics[width=\linewidth]{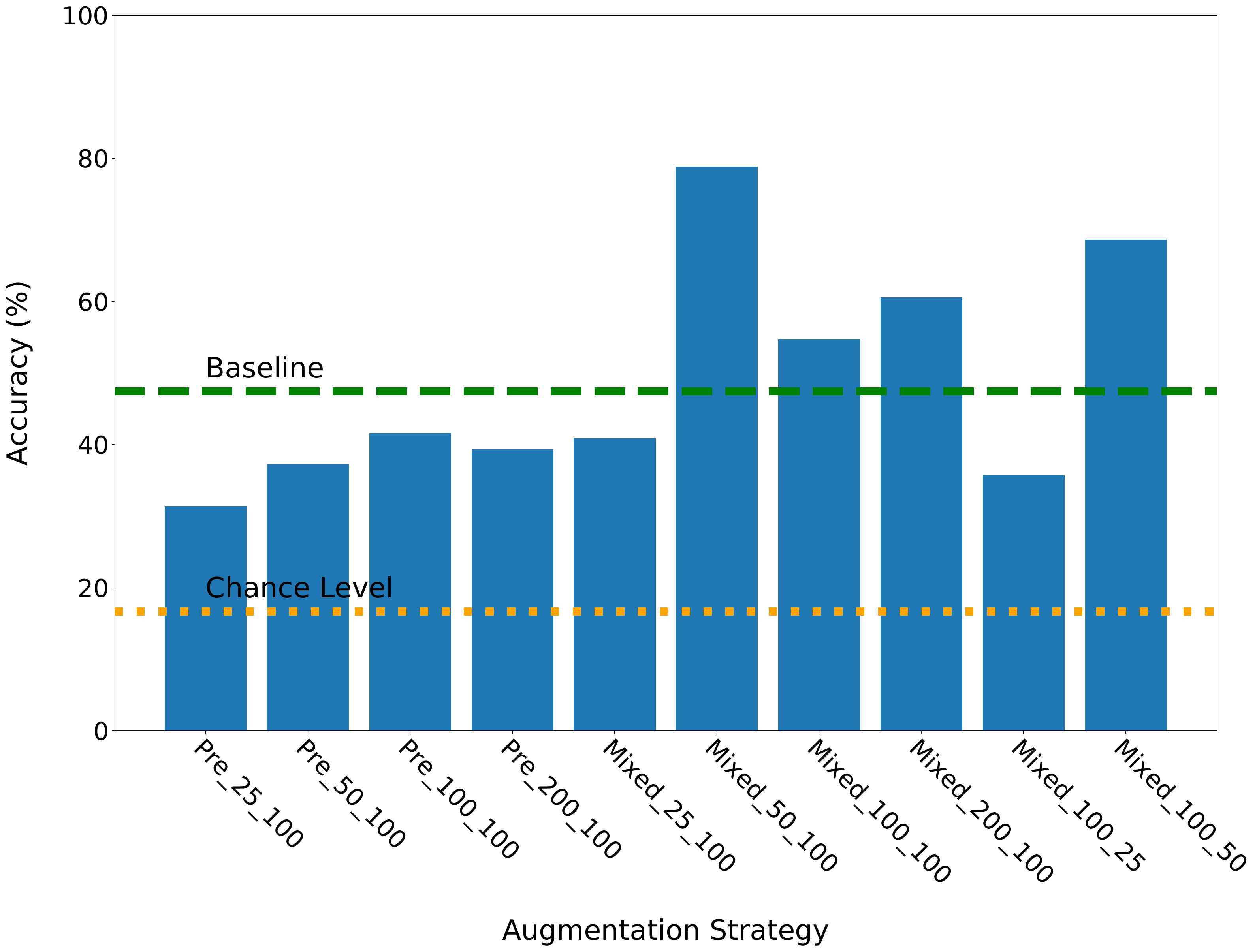}
\caption{Gesture recognition accuracy for different combination of synthetic and real hand gestures. The prefix ``\textit{Pre}'' are the models pre-trained with the synthetic data, while ``\textit{Mixed}'' are models trained from scratch with the combined synthetic and real data. The first number is the percentage of gesture variations (i.e., data size) for synthetically generated hand gestures, while the second number is the same for the real data.} 
\label{fig:results}
\end{figure}

A comprehensive set of experiments was performed to evaluate the effectiveness of synthetic data and to determine optimal configurations for training a neural network. The baseline experiment exclusively utilized real data, consisting of a diverse range of 33 gesture variations per gesture class collected from 20 participants~\cite{molchanov2016online}. To assess the efficacy of synthetic data, a pre-training approach was used, where different proportions of synthetic data were initially trained, corresponding to ratios of 25\%, 50\%, 100\% and 200\% (i.e., 8, 16, 33, and 66 generated gesture variations per gesture). Then, it was followed by a fine-tuning step using the entire real dataset. Subsequent experiments involved the training of the neural network from scratch using this combination of synthetic and real data throughout the training process, with the hyperparameters set to default values as defined in~\cite{kopuklu2019real}. Consistency in training time and epoch count was maintained to ensure a valid comparison of the trained models.
Furthermore, two additional experiments were carried out, combining 25\% and 50\% of the real data variations with 100\% of the synthetic data variations. These experiments involved simultaneous training in both real and synthetic data, replicating the methodology mentioned earlier to examine the individual impact of real and synthetic data on the learning model. Figure~\ref{fig:results} presents the recognition accuracy for all experiments. It is evident that the pretraining strategy did not enhance accuracy compared to the baseline, indicating that using synthetic data alone is not equivalent to using real data. On the contrary, training the model from scratch with a combination of synthetic and real data demonstrated superior performance, particularly when a substantial amount of synthetic data was incorporated.
Furthermore, when comparing the ``\textit{Mixed\_25\_100}'' model to the ``\textit{Mixed\_100\_25}'' model, with accuracies of $40.87\%$ and $35.76\%$, respectively, both synthetic and real data showed a similar effect on recognition performance. Similarly, comparing the ``\textit{Mixed\_50\_100}'' model to the ``\textit{Mixed\_100\_50}'' model, with accuracies of $78.83\%$ and $68.61\%$, respectively, revealed a similar pattern. However, in both cases, having a higher proportion of real data than synthetic data resulted in improved performance.
These experiments highlight that models incorporating synthetic with real data outperform those trained solely on real data, specifically those trained with a substantial amount of synthetic data.

Furthermore, we analyze the impact of different ranges of variations, including speed, position, and finger spacing, on recognition performance, as in related work~\cite{lindgren2018learned}. Each variation is examined under median, low, and high range conditions. We chose the model ``\textit{Mixed\_100\_100}'' as the default model for this analysis due to its high accuracy and moderate complexity.
In the low-range condition, gestures are generated within half of the median range, while in the high-range condition, gestures are generated within double the median range. For example, if the median range for the speed variation adds values between 0 and 50 cm/s to the default speed, the low-range condition will have a range of 12.5 to 37.5 cm/s, and the high-range condition will have a range of -25 to 75 cm/s. This assumes that the default speed is above 25 cm/s to ensure positive values. The low-range and high-range conditions for position and finger spacing variations follow a similar pattern, with each dimension or finger having its own adjusted range.
It is important to note that only one variation's range is altered at a time, whereas all other variations, including other active variations besides speed, position, and finger spacing, continue to generate random values within their respective median ranges. This approach maintains a controlled number of changing variables, which is desirable to accurately interpret the experimental results. Restricting the variations exclusively to finger spacing, for example, while keeping all other parameters constant, would result in repeated gestures with minimal differences in finger positions, rendering the analysis less meaningful. 

\begin{table*}[t]
\begin{center}

\resizebox{0.85\textwidth}{!}{
\begin{tabular}{|l|c|c|c|c|c|}
\hline
& \multicolumn{3}{|c|}{Gesture-Related}  & 
    			\multicolumn{2}{|c|}{Camera-Related}\\
    			\hline
Range Condition & Speed & Position & Finger Spacing & Chromaticity & Depth Range \\
\hline\hline
Low & 40.87\% & 43.06\% & 41.60\% & 42.33\% &  40.87\% \\
Median & 54.74\% & \textbf{54.74\%}  & \textbf{54.74\%}  & \textbf{54.74\%}  & \textbf{54.74\%}  \\
High & \textbf{55.47\%}  & 40.87\%  &  31.38\% &  50.36\% & 41.60\% \\
\hline
\end{tabular}
}
\end{center}
\caption{Comparison of accuracy performance across different ranges of varied gesture-related and camera-related parameters. Note that the median range value remains consistent for all variations, as it represents the accuracy result obtained from the default model ``\textit{Mixed\_100\_100}''.}
\label{tab:variationresults}
\end{table*}

Similarly to the previous approach, an analysis of other parameters of camera settings is conducted. As evaluation involves a depth camera, a similar conditional analysis is applied to certain modeling aspects of the camera, such as the linear coefficient of the camera chromaticity values~\cite{van2018hue} and the minimum and maximum range of the camera, which typically represents a hardware specification for various commercially available depth cameras. The comparison of the model performance for gesture-related and camera-related parameters under the three range conditions is presented in~\autoref{tab:variationresults}. Accuracy results demonstrate that the selection of appropriate parameters and settings for gesture generation significantly improves the quality of synthetic gestures. For example, excessively limiting the range of variations (i.e., low-range condition) leads to overfitting and decreases performance, while excessively high variance (i.e., high-range condition) introduces additional noise. However, modeling both low and high ranges is crucial to align with realistic hardware capabilities and gesture performance constraints. It is worth mentioning that the median range consistently yields optimal results across all ranges, as it is modeled to represent the most realistic settings based on previous research and assumes a highly capable hardware configuration.

\section{Discussion and Conclusion}

We introduce SynthoGestures, a novel framework for generating synthetic dynamic hand gestures to enhance the accuracy of gesture recognition models. By leveraging the power of virtual 3D models and animation software, our framework enables the synthesis of diverse and customizable gesture datasets. We have shown that by incorporating variations such as gesture speed, performance, and hand shape, along with simulating multiple camera locations and types, our framework produces natural-looking gestures that closely resemble real-world scenarios.
Furthermore, our experiments have shown that SynthoGestures can effectively augment existing real-hand datasets while enhancing their performance. This capability saves significant time and effort in the creation of datasets and accelerates the development of gesture recognition systems in automotive applications and beyond.
While the results have demonstrated the effectiveness of SynthoGestures in improving gesture recognition performance, it is essential to acknowledge certain limitations and potential avenues for future work. First, the current framework focuses primarily on the visual aspect of hand gestures and may not capture other sensory cues, such as tactile or proprioceptive feedback. For example, trimmers could be added as noise to the hand gestures to simulate force feedback when driving on different terrains. Exploring ways to incorporate these additional modalities could further enhance the authenticity and realism of synthesized gestures. Moreover, although SynthoGestures provides a wide range of variations in gesture speed, performance, and hand shape, there is still room to explore more complex and nuanced variations that can capture the intricacies of natural hand movements. Finally, our results show that having a higher ratio of real-to-synthetic data is essential for improving the model training performance, which should be considered in future work and investigated further.

In conclusion, our framework can facilitate future work, advance the field of static and dynamic hand gesture recognition, and pave the way for more sophisticated and accurate human-computer interaction systems in various domains, including automotive interfaces, virtual reality, and augmented reality applications.

% These future directions hold great promise for revolutionizing the intersection of robotics, artificial intelligence, and healthcare.

%===============================================================================

\addtolength{\textheight}{-14cm}   % This command serves to balance the column lengths
                                  % on the last page of the document manually. It shortens
                                  % the textheight of the last page by a suitable amount.
                                  % This command does not take effect until the next page
                                  % so it should come on the page before the last. Make
                                  % sure that you do not shorten the textheight too much.

%%%%%%%%%%%%%%%%%%%%%%%%%%%%%%%%%%%%%%%%%%%%%%%%%%%%%%%%%%%%%%%%%%%%%%%%%%%%%%%%

%%%%%%%%%%%%%%%%%%%%%%%%%%%%%%%%%%%%%%%%%%%%%%%%%%%%%%%%%%%%%%%%%%%%%%%%%%%%%%%%

%%%%%%%%%%%%%%%%%%%%%%%%%%%%%%%%%%%%%%%%%%%%%%%%%%%%%%%%%%%%%%%%%%%%%%%%%%%%%%%%
% \section*{APPENDIX}

% Appendixes should appear before the acknowledgment.

\section*{ACKNOWLEDGMENT}

This work is partially funded by the German Ministry of Education and Research (BMBF) under the TeachTAM project (Grant Number: 01IS17043), the CAMELOT project (Grant Number: 01IW20008) and the FedWell project (Grant Number: 01IW23004).

%%%%%%%%%%%%%%%%%%%%%%%%%%%%%%%%%%%%%%%%%%%%%%%%%%%%%%%%%%%%%%%%%%%%%%%%%%%%%%%%

\bibliographystyle{IEEEtran}
\bibliography{IEEEabrv,IEEEexample}

\end{document}